\documentclass[conference]{IEEEtran}
\IEEEoverridecommandlockouts

\usepackage{cite}           % IEEE-style citation handling (replaces natbib from the NeurIPS style)
\usepackage[utf8]{inputenc} % allow utf-8 input
\usepackage[T1]{fontenc}    % use 8-bit T1 fonts
\usepackage{url}            % simple URL typesetting
\usepackage{booktabs}       % professional-quality tables
\usepackage{amsfonts}       % blackboard math symbols
\usepackage{amssymb}        % \blacktriangledown etc.
\usepackage{nicefrac}       % compact symbols for 1/2, etc.
\usepackage{microtype}      % microtypography
\usepackage{xcolor}         % colors
\usepackage{graphicx}
\usepackage{amsmath}
\usepackage{tikz}
\usetikzlibrary{arrows.meta,positioning,shapes.geometric,shapes.symbols}

\usepackage{multirow}
\usepackage{makecell}
\usepackage[font=small,labelfont=bf]{caption}
\usepackage{subcaption}
\usepackage[normalem]{ulem}  % \uline for line-breakable underline (RQ answers at end of subsection)
\usepackage{comment}         % \begin{comment} ... \end{comment} for block comments
\usepackage{xr-hyper}        % cross-document \ref (main <-> appendix split); load BEFORE hyperref
\usepackage{hyperref}        % hyperlinks
\begin{document}

\title{Seeking Flat Minima: Robust Data Reshaping via Perturbation-Aware Diffusion and Trajectory Ensembles}
\title{Robust Data Reshaping via Perturbation-Aware Diffusion and Trajectory Ensembles}
\title{Robust Data Reshaping via Flat and Consensus-Seeking Diffusion}
\title{Flat-Consensus Diffusion for Robust Data Reshaping under Noisy Evaluator}
%\title{Flat-Consensus Diffusion for Robust Data Reshaping}

% Fill in the author block for the camera-ready / submission as needed.
% IEEE conference author format:
% Author block (order: Hongyu Cao, Yanjie Fu, Kunpeng Liu).

\author{%
  \IEEEauthorblockN{Hongyu Cao}
  \IEEEauthorblockA{Clemson University \\
    Clemson, SC, USA \\
    hcao2@clemson.edu}
  \and
  \IEEEauthorblockN{Kunpeng Liu}
  \IEEEauthorblockA{Clemson University \\
    Clemson, SC, USA \\
    kunpenl@clemson.edu}
  \and
  \IEEEauthorblockN{Fei Xie}
  \IEEEauthorblockA{Portland State University \\
    Portland, OR, USA \\
    xie@pdx.edu}
  \and
  \IEEEauthorblockN{Sandip Ray}
  \IEEEauthorblockA{University of Florida \\
    Gainesville, FL, USA \\
    sandip@ece.ufl.edu}
}

\maketitle

% TODO：需要考虑把论文的叙述改成ensemble学习解决uncertain evaluator

%%%%%%%%%%%%%%%%%%%%%%%%%%%%%%%%%%%%%%%%%%%%%%%%%%%%%%%%
% ========== Data & Distribution ======================
%%%%%%%%%%%%%%%%%%%%%%%%%%%%%%%%%%%%%%%%%%%%%%%%%%%%%%%%

% empirical datasets
\newcommand{\traindataX}{\mathbf{X}_{\mathrm{train}}}
\newcommand{\valdataX}{\mathbf{X}_{\mathrm{val}}}
\newcommand{\testdataX}{\mathbf{X}_{\mathrm{test}}}

\newcommand{\traindatay}{\mathbf{y}_{\mathrm{train}}}
\newcommand{\valdatay}{\mathbf{y}_{\mathrm{val}}}
\newcommand{\testdatay}{\mathbf{y}_{\mathrm{test}}}

% underlying data distributions
\newcommand{\trainDist}{P_{\mathrm{train}}}
\newcommand{\valDist}{P_{\mathrm{val}}}
\newcommand{\testDist}{P_{\mathrm{test}}}

% feature space (column schema)
\newcommand{\featurespace}{\mathcal{F}}

%%%%%%%%%%%%%%%%%%%%%%%%%%%%%%%%%%%%%%%%%%%%%%%%%%%%%%%%
% ========== Feature Operations =======================
%%%%%%%%%%%%%%%%%%%%%%%%%%%%%%%%%%%%%%%%%%%%%%%%%%%%%%%%

% primitive operation set
\newcommand{\operationset}{\mathcal{O}}

% symbolic operation sequence (grammar-level object)
\newcommand{\opseq}{\mathbf{o}}

% executable transformation induced by opseq
\newcommand{\transform}{T}

% transformation application operator
\newcommand{\applytransform}{\mathcal{A}}

% decoder: latent -> transformation
\newcommand{\decoder}{D}

%%%%%%%%%%%%%%%%%%%%%%%%%%%%%%%%%%%%%%%%%%%%%%%%%%%%%%%%
% ========== Latent Space =============================
%%%%%%%%%%%%%%%%%%%%%%%%%%%%%%%%%%%%%%%%%%%%%%%%%%%%%%%%

\newcommand{\latent}{z}
\newcommand{\latentspace}{\mathcal{Z}}

%%%%%%%%%%%%%%%%%%%%%%%%%%%%%%%%%%%%%%%%%%%%%%%%%%%%%%%%
% ========== Downstream Model =========================
%%%%%%%%%%%%%%%%%%%%%%%%%%%%%%%%%%%%%%%%%%%%%%%%%%%%%%%%

\newcommand{\downstream}{\mathcal{M}}

%%%%%%%%%%%%%%%%%%%%%%%%%%%%%%%%%%%%%%%%%%%%%%%%%%%%%%%%
% ========== Reward Definitions =======================
%%%%%%%%%%%%%%%%%%%%%%%%%%%%%%%%%%%%%%%%%%%%%%%%%%%%%%%%

% true population reward
\newcommand{\truereward}{R^{\ast}}

% train-distribution reward
\newcommand{\trainreward}{R_{\mathrm{train}}}

% test-distribution reward
\newcommand{\testreward}{R_{\mathrm{test}}}

% evaluator approximation
\newcommand{\evaluator}{\widehat{R}}

% robust reward
\newcommand{\robustreward}{R_{\mathrm{robust}}}

%%%%%%%%%%%%%%%%%%%%%%%%%%%%%%%%%%%%%%%%%%%%%%%%%%%%%%%%
% ========== Robustness / Perturbation ================
%%%%%%%%%%%%%%%%%%%%%%%%%%%%%%%%%%%%%%%%%%%%%%%%%%%%%%%%

\newcommand{\perturb}{\xi}
\newcommand{\perturbset}{\Delta}

% aggregation operator (trimmed mean, median, etc.)
\newcommand{\agg}{\mathrm{Agg}}

%%%%%%%%%%%%%%%%%%%%%%%%%%%%%%%%%%%%%%%%%%%%%%%%%%%%%%%%
% ========== Loss & Utility ===========================
%%%%%%%%%%%%%%%%%%%%%%%%%%%%%%%%%%%%%%%%%%%%%%%%%%%%%%%%

\newcommand{\loss}{\ell}
\newcommand{\utility}{U}

%%%%%%%%%%%%%%%%%%%%%%%%%%%%%%%%%%%%%%%%%%%%%%%%%%%%%%%%
% ========== Problem-Specific Symbols =================
%%%%%%%%%%%%%%%%%%%%%%%%%%%%%%%%%%%%%%%%%%%%%%%%%%%%%%%%

% raw feature matrix
\newcommand{\rawdata}{\mathbf{X}}

% target vector
\newcommand{\rawtarget}{\mathbf{y}}

% downstream evaluator
\newcommand{\evaluatorfunc}{\mathcal{E}}

% reshaping policy (operation sequence)
\newcommand{\policy}{S}

% policy length
\newcommand{\policylen}{L}

% operation element
\newcommand{\operation}{o}

% transformation induced by policy
\newcommand{\policytransform}{\phi_{\policy}}

% generative search policy
\newcommand{\genpolicy}{\pi_{\theta}}

% policy parameter
\newcommand{\policyparam}{\theta}

% noise distribution
\newcommand{\noisedist}{\mathcal{D}_{\mathrm{noise}}}

% optimal policy parameter
\newcommand{\optimalpolicyparam}{\theta^{\ast}}

% optimal reshaping policy
\newcommand{\optimalpolicy}{S^{\ast}}

\newcommand{\methodname}{FCDiff}

\begin{abstract}
Data shape determines how features are structured, how patterns are separated, and how distributions cover the underlying domain. Poor data shape can make models learn noise rather than generalizable structure. This paper studies robust feature-centric data reshaping: generating feature transformations that remain useful, stable, and reproducible under noisy evaluation and imperfect data conditions. We view reshaping operation sequence search as reward-guided diffusion generation, and robust reshaping as searching for regions in the latent reward landscape rather than isolated high-reward transformations. The key challenge is dual instability: noisy evaluators distort local reward guidance, while stochastic generative trajectories can converge to inconsistent solutions. We propose \methodname{}, a flat-consensus diffusion framework that addresses both failures through a micro--macro decomposition. The micro layer replaces point-estimate reward guidance with Gaussian-smoothed, Monte Carlo averaged gradients, steering generation toward locally flat reward regions. The macro layer aggregates independently guided trajectories with a weighted Fr\'echet-mean barycenter, selecting consensus-supported basins and filtering stochastic outliers. Across an 8-dataset headline cohort under heavy-tailed evaluator noise, \methodname{} attains the best aggregate rank on lower-tail reliability and robustness against both search-based AutoFE and robustness-oriented generative baselines, with statistically significant accuracy gains over every generative baseline. 
%Our results show that robust data reshaping is not achieved by selecting a single noisy optimum, but by finding transformations in regions that are both locally flat and globally reproducible.
Our results show that robust data reshaping requires searching for flat, consensus-supported regions rather than sharp single-trajectory optima.  
\end{abstract}

\section{Introduction}

Real-world data rarely arrives in a shape that matches downstream learning tasks. In healthcare, materials science, transportation, energy, fraud detection, and cybersecurity, raw data is often noisy, sparse, imbalanced, and weakly separated. In these settings, model quality depends not only on model architecture or data scale, but also on data shape: how features are structured, how patterns are separated, and how distributions cover the domain.

Feature-centric data reshaping aims to improve data shape by transforming and generating features, and has become a key tool for data-centric learning~\cite{khurana2018feature,chen2019neural,ling2025gradient}. Concretely, a reshaping method applies a sequence of operations to the feature table, such as taking $\log(\text{income})$ or crossing age with BMI, and produces a new feature table. A downstream model is then trained on the new table, and its validation performance serves as the reward that guides the search for better operation sequences. However, most methods are designed for clean evaluation settings, while real-world reshaping must operate under noisy downstream feedback, perturbed measurements, and unstable search dynamics. This paper studies robust feature-centric data reshaping: how to generate feature transformations that remain useful, stable, and reproducible under imperfect data conditions.

\begin{figure}[!t]
    \centering
    \includegraphics[width=\columnwidth]{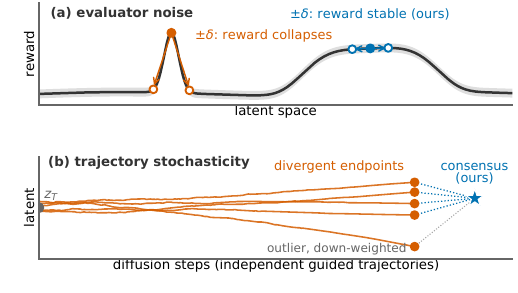}
    \caption{Robust data reshaping at a glance: two coupled instabilities. \textbf{(a)} Evaluator noise. Search runs over a latent reward landscape scored by a noisy evaluator (gray band); under a small latent perturbation $\pm\delta$, a sharp peak loses its reward while a flat region keeps it. \textbf{(b)} Trajectory stochasticity. Independent guided trajectories from the same noise $z_T$ end at different latents; we aggregate them into a weighted consensus (star) that down-weights stochastic outliers. \methodname{} targets regions that are both flat and consensus-supported.}
    \label{fig:teaser}
\end{figure}

The core difficulty is that robust reshaping must handle two coupled sources of instability. First, noisy evaluators distort reward-guided search: data corruption, model sensitivity, and cross-validation randomness can make a brittle transformation appear useful, causing the search to overfit reward artifacts. Second, stochastic generative search is inherently unstable. Each search run follows its own random path, which we call a trajectory. Different trajectories may converge to different local optima, even when guided by the same reward model. These failures reinforce each other. A method that only smooths reward noise may still produce inconsistent trajectories, while a method that only aggregates trajectories may average around noisy sharp optima. Robust reshaping therefore requires both local stability under perturbation and global agreement across trajectories.

Existing work only partially addresses this dual-instability problem. Reinforcement learning and evolutionary AutoFE methods search transformation sequences using downstream reward, but treat noisy finite-sample feedback as reliable and therefore overfit reward artifacts~\cite{khurana2018feature,chen2019neural}. Gradient-based and surrogate-based reshaping methods improve optimization efficiency, but still follow single optimization trajectories and do not address stochastic search instability~\cite{ling2025gradient, liu2026continuousoptimizationfeatureselection,liu2026hierarchicalpermutationinvariantfeaturetransformation}. Generative and diffusion-based methods offer broader exploration of large transformation spaces, but typically rely on point-estimate reward gradients and lack cross-trajectory consensus mechanisms~\cite{kim2024diffusion}. In short, existing methods address search, reward, or generation in isolation, but not the joint problem of noise-robust guidance and trajectory-level reproducibility.

Our central perspective is simple: optimal data reshaping action search is reward-guided diffusion generation, and robust data reshaping is robustness-guided generation. In this view, discrete operation sequences are embedded in a continuous latent space, and each latent point has a predicted reward. This reframes robustness as a geometric property of this latent reward landscape. Robust reshaping actions should not lie on sharp single-trajectory reward peaks; they should lie in regions that are both locally flat under perturbation and consistently reached by multiple trajectories. This yields two design principles: within-trajectory perturbation insensitivity and between-trajectory agreement. Based on this view, we make three contributions. First, we formulate robust feature-centric reshaping as a dual-instability problem involving evaluator noise and trajectory stochasticity. Second, we develop a dual-level generative solution that explicitly targets both failure modes. Third, we empirically show that robust reshaping succeeds when it searches for flat, consensus-supported regions rather than optimizing a single noisy point estimate.

We propose \methodname{}, a diffusion framework for robust data reshaping that searches for flat, consensus-supported regions. At the micro level, \methodname{} uses Gaussian smoothing and Monte Carlo averaged gradients~\cite{cohen2019certified,duchi2012randomized} to replace point-estimate reward guidance, steering generation toward flat reward regions. At the macro level, \methodname{} uses a weighted Fr\'echet-mean barycenter~\cite{ganaie2022ensemble} to aggregate multiple guided trajectories into a consensus latent, filtering stochastic outliers while preserving lower-tail reward. These mechanisms are complementary: the micro layer reduces sensitivity to local reward noise, while the macro layer improves reproducibility across stochastic runs. Across 8 datasets under heavy-tailed evaluator noise, \methodname{} achieves the strongest lower-tail reliability, robustness, and reproducibility compared with AutoFE search and robustness-oriented generative baselines. The key finding is that robust data reshaping is not about finding the highest single reward peak; it is about finding feature transformations in flat, consensus-supported regions.
Robust data reshaping reduces to a two-dimensional problem: local flatness ensures stability under evaluator perturbations, while global consensus ensures reproducibility under stochastic search.

\section{Problem Statement}
%We formalize feature-centric data reshaping as a search problem over feature transformation sequences.

\paragraph{Feature-Centric Data Reshaping.}
Let $\rawdata \in \mathbb{R}^{N \times F}$ be the input feature matrix and $\rawtarget \in \mathbb{R}^{N}$ be the prediction targets, $N$ is the number of instances, $F$ is the number of features, $\operationset$ is a library of atomic feature transformation operations like logarithmic, interaction, and discretization . 
A reshaping operation sequence $\policy = [\operation_1, \dots, \operation_{\policylen}]$, with $\operation_l \in \operationset$, maps the raw feature space to a transformed feature space $\mathcal{X}' = \policytransform(\rawdata)$. A downstream evaluator assigns a reward score $\evaluatorfunc(\policytransform(\rawdata))$ based on predictive utility~\cite{cast_norm}. 
The goal is to search reshaping operation sequences that optimize this reward.

\paragraph{Two Instabilities.}
%Real-world reshaping faces two coupled instabilities. 
First, \emph{reward-level instability} arises because the evaluator is noisy: data corruption, model hyperparameter sensitivity, and cross-validation randomness can make the same sequence receive inconsistent rewards. We denote such perturbations by $\perturb$, and write $\evaluatorfunc_{\perturb}$ for the evaluator realization under $\perturb$.
Second, \emph{trajectory-level instability} arises because generative search is stochastic: multiple independent search trajectories, each of which is denoted by $\tau$, can converge to different instable (high-reward but inconsistent) reshaping operation sequences. Thus, a robust method must avoid both noisy sharp optima and seed-dependent solutions.

\paragraph{The Robust Reshaping Objective.}
We aim to discover the optimal generative policy $\genpolicy$ that produces a reshaping operation sequence $S$ with high utility, local perturbation stability, and cross-trajectory consensus:
\begin{equation}
\label{eq:robust-objective}
\optimalpolicyparam ~=~ \operatorname*{arg\,max}_{\policyparam}~
\mathbb{E}_{\tau \sim \genpolicy}\,
\mathbb{E}_{\perturb}
\big[\,\evaluatorfunc_{\perturb}\big(\phi_{\policy(\tau)}(\rawdata)\big)\,\big],
\end{equation}
where $\policy(\tau)$ is the reshaping sequence produced by trajectory $\tau$.
% This objective captures both robustness to evaluator perturbations and stability across stochastic search trajectories.

\paragraph{Design Implication.}
The formulation implies: i) within each trajectory, reward guidance should be insensitive to local perturbations; ii) across trajectories, the final solution should be supported by consensus rather than selected from a single stochastic run. Section~\ref{sec:methodology} instantiates these requirements with micro-level \emph{randomized smoothing} and macro-level \emph{weighted Fr\'echet-mean} aggregation.

\section{Robust Data Reshaping}
\label{sec:methodology}

\begin{figure*}[!t]
    \centering
    \includegraphics[width=0.9\textwidth]{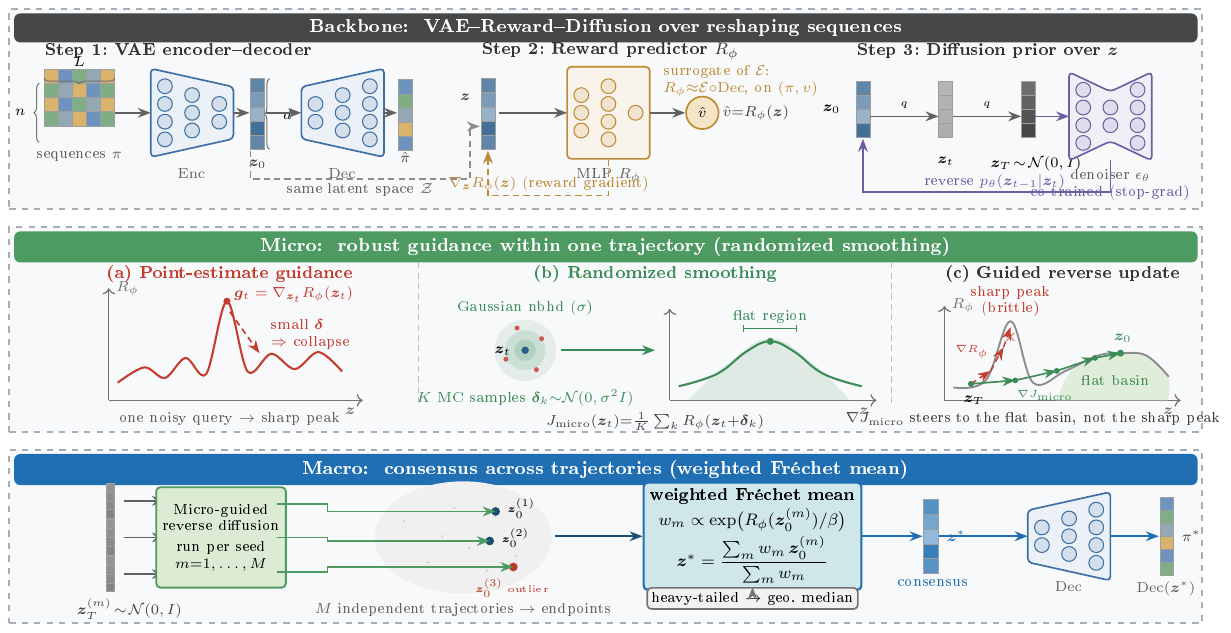}
    \caption{
    FCDiff as a symmetric micro--macro framework for robust data reshaping.
    A VAE--reward--diffusion backbone builds a continuous latent reward geometry over discrete reshaping sequences.
    The micro layer handles local reward instability by smoothing point-estimate guidance into flat-region guidance.
    The macro layer handles trajectory-level instability by aggregating multiple guided endpoints into a consensus latent.
    Together, FCDiff searches for flat, consensus-supported regions rather than sharp single-trajectory optima.
    }
    \label{fig:overview}
\end{figure*}
We take a generative perspective: reshaping is not the problem of finding one optimal sequence, but of identifying a region in the transformation space that is both locally stable and consistently reachable. This reframes robust reshaping as searching for flat, consensus-supported regions in a stochastic reward landscape, rather than maximizing a single noisy objective. We propose \textbf{FCDiff}, a diffusion framework for robust data reshaping that searches for flat, consensus-supported regions rather than sharp single-trajectory optima. Guided diffusion provides the key substrate: a continuous latent space in which local smoothing and global aggregation can be applied jointly during generation. At the \emph{micro level}, FCDiff replaces point-estimate reward guidance with Gaussian-smoothed gradients, biasing generation toward flat regions that are insensitive to evaluator perturbations. At the \emph{macro level}, FCDiff replaces single-trajectory selection with trajectory aggregation, identifying solutions that lie in consensus-supported basins rather than isolated optima. This enables robustness-aware search beyond greedy or single-trajectory methods.

\subsection{Backbone: Latent Reward Geometry for Generative Reshaping}
\label{sec:backbone}

Robust reshaping requires searching for regions that are locally flat under perturbations and consistently reachable across trajectories. This is difficult in the original discrete transformation space, where search is combinatorial and rewards are sparse, noisy, and non-differentiable. We therefore adopt the VAE--reward--diffusion backbone of DIFFT~\cite{diffusionfeature2025}, on top of which our two ensemble layers (\S\ref{sec:micro}, \S\ref{sec:macro}) are built; the \texttt{Std-Diff} row of Table~\ref{tab:main_2026_04_22} is exactly this vanilla DIFFT backbone with $K{=}M{=}1$. This backbone instantiates the generative policy $\genpolicy$ in Eq.~\eqref{eq:robust-objective}: the VAE defines the continuous search space, the reward-guided diffusion prior realizes stochastic sequence generation, and the decoder maps sampled latents back to reshaping sequences $\policy$.

First, a VAE maps each discrete reshaping sequence $\policy$ to a latent code $z_0 = \mathrm{Enc}(\policy)$, with $\policy \approx \mathrm{Dec}(z_0)$, enabling gradient-based search while preserving decode validity. Second, a differentiable reward predictor $R_\phi: \mathcal{Z} \to \mathbb{R}$, a surrogate of the downstream evaluator $\evaluatorfunc$, is jointly trained on sequence--reward pairs, shaping the latent space into a reward-aware geometry where nearby points have meaningful utility structure; in this geometry, flat regions correspond to reshaping strategies whose predicted utility is stable under perturbation. Third, a diffusion prior~\cite{ho2020denoising} models the distribution of valid latent sequences by progressively noising $z_0$ to $z_T \approx \mathcal{N}(0,I)$ and learning the reverse denoising process. This gives FCDiff a stochastic generative search process over valid transformations.

The backbone is not the robustness mechanism itself; it provides the framework on which robustness can be imposed. The VAE supplies a smooth search space, the reward predictor supplies a utility-aligned geometry, and diffusion supplies stochastic trajectories, enabling local perturbation smoothing (\S\ref{sec:micro}) and cross-trajectory consensus (\S\ref{sec:macro}).

\subsection{Micro: Local Stability via Randomized Smoothing}
\label{sec:micro}

Point-estimate reward guidance is brittle under noisy evaluators. In our setting, evaluator perturbations $\perturb$ enter the search as stochastic reward queries, i.e., as noise on the guidance gradient $\nabla_{z_t} R_\phi(z_t)$; latent-space flatness is therefore the search-space counterpart of evaluator-perturbation stability --- a correspondence verified by the gradient-variance and reward-floor diagnostics in Sec.~\ref{sec:experiments}, and one that holds insofar as the noise is variance-like rather than a systematic surrogate bias. During reverse diffusion, a single noisy gradient can pull the trajectory toward a sharp latent point that appears high-reward under one evaluator realization but fails under small perturbations. This is the within-trajectory instability in Eq.~\eqref{eq:robust-objective}: the search overfits local reward artifacts.

FCDiff addresses this by replacing point-estimate guidance with local smoothing~\cite{cohen2019certified,duchi2012randomized}. At each reverse step $t$, instead of using $\nabla_{z_t} R_\phi(z_t)$, we evaluate the reward predictor over a Gaussian neighborhood around $z_t$:
\begin{equation}
\label{eq:smooth_obj}
J_{\mathrm{micro}}(z_t) \;=\; \frac{1}{K}\sum_{k=1}^{K} R_\phi(z_t + \delta_k), \qquad \delta_k \sim \mathcal{N}(0,\sigma^2 I).
\end{equation}
where $K$ is the number of MC perturbations, $\delta_k$ are i.i.d.~Gaussian noise with smoothing radius $\sigma$ (tied to the per-dimension latent scale), and $J_{\mathrm{micro}}$ is the locally smoothed surrogate of $R_\phi$.

The reverse update then uses $\nabla_{z_t} J_{\mathrm{micro}}(z_t)$ as the guidance direction. This changes the guidance signal from attraction to a single point into attraction to a locally stable region.

Geometrically, randomized smoothing favors latent regions where nearby perturbations preserve high reward. Thus, one guided trajectory produces an endpoint $z_0$ whose predicted utility is less sensitive to evaluator noise. This implements the first robustness requirement: local perturbation stability within a single diffusion trajectory.

\subsection{Macro: Global Consistency via Trajectory Consensus}
\label{sec:macro}

Local smoothing stabilizes each trajectory, but stochastic generation can still produce divergent endpoints across runs. Even when each trajectory is locally robust, different noise realizations of the reverse diffusion process may converge to different high-reward but inconsistent reshaping strategies. This is the between-trajectory instability in Eq.~\eqref{eq:robust-objective}: the final solution depends on a single stochastic realization.

FCDiff addresses this by replacing single-trajectory selection with trajectory consensus. We run $M$ independently guided trajectories and obtain endpoints $\{z_0^{(1)}, \dots, z_0^{(M)}\}$. Instead of selecting the highest-reward sample, we compute a weighted Fr\'echet mean:
\begin{equation}
\label{eq:frechet}
z^* \;=\; \operatorname*{arg\,min}_{z} \sum_{m=1}^{M} w_m \, \|z - z_0^{(m)}\|^2,
\end{equation}
where $z^*$ is the consensus latent aggregating the $M$ endpoints into a single representative point, with weights
\begin{equation}
\label{eq:softmax_w}
w_m \;\propto\; \exp\!\big(R_\phi(z_0^{(m)})/\beta\big).
\end{equation}
where $\beta>0$ is a softmax temperature and $R_\phi$ is the surrogate, not the downstream evaluator $\evaluatorfunc$; $\beta\!\to\!\infty$ recovers the unweighted Fr\'echet mean, $\beta\!\to\!0^+$ collapses to top-1 selection. Eq.~\eqref{eq:frechet} admits the closed form $z^* = \sum_{m=1}^{M} w_m z_0^{(m)} \big/ \sum_{m=1}^{M} w_m$, i.e., a reward-weighted average of the trajectory endpoints.

This changes the selection rule from ``best sample'' to ``center of agreement.''

Geometrically, robust reshaping strategies form broad basins that are repeatedly reached across trajectories, while spurious solutions appear as isolated peaks visited by only a few runs. The weighted Fr\'echet mean concentrates on high-reward, high-consensus regions and filters trajectory-level outliers. Decoding $z^*$ gives the final reshaping sequence $\optimalpolicy = \mathrm{Dec}(z^*)$. This implements the second robustness requirement: global reproducibility across diffusion trajectories.
The idea that a single level of consistency is insufficient also appears in
generative imputation, where multi-level causal constraints outperform
single-level ones~\cite{GIMCC}.

\paragraph{Micro--Macro Symmetry.}
The micro and macro components of FCDiff form a symmetric decomposition of robust reshaping: the micro layer stabilizes guidance within a single trajectory, while the macro layer stabilizes selection across trajectories. Table~\ref{tab:micro-macro} summarizes this correspondence.

\begin{table}[t]
\centering
\footnotesize
\setlength{\tabcolsep}{2.5pt}
\begin{tabular}{@{}lcc@{}}
\toprule
 & \textbf{Micro (Local)} & \textbf{Macro (Global)} \\
\midrule
Failure Mode & Noisy reward gradients & \makecell{Stochastic trajectory\\divergence} \\
Operator & Gaussian smoothing & Fr\'echet mean aggregation \\
Objective Relaxation & $R(z) \rightarrow \mathbb{E}_{\delta}[R(z+\delta)]$ & $\max \rightarrow$ consensus center \\
Geometry & Flat regions & Broad basins \\
Effect & Reduces local sensitivity & Reduces cross-run variance \\
Output & Stable trajectory endpoint & Consensus latent solution \\
\bottomrule
\end{tabular}
\caption{Structural symmetry of the micro and macro components in FCDiff. The two layers address complementary sources of instability and jointly enable region-level robust reshaping.}
\label{tab:micro-macro}
\end{table}

Taken together, FCDiff turns robust data reshaping from point-wise optimization into region-level inference over a structured latent space. The backbone constructs a continuous reward-aware geometry and a stochastic trajectory distribution; the micro layer enforces local flatness through randomized smoothing; and the macro layer enforces global consensus through trajectory aggregation. These components form a coordinated system for addressing the two instability sources identified in Section~2. Section~4 empirically verifies that this micro--macro decomposition improves reliability, robustness, and reproducibility under noisy evaluators.
\section{Experiments}
\label{sec:experiments}

Our experiments test the central claim of \methodname{}: robust data reshaping requires both local flatness and global consensus. 
We evaluate downstream reliability under noisy evaluators, local stability from randomized smoothing, and cross-trajectory reproducibility from consensus aggregation. 
All methods use the same noisy-evaluator protocol: corrupted reward queries during search and clean downstream evaluation after search. 
We report both performance metrics and mechanism diagnostics to show not only whether \methodname{} works, but also why it works.

\paragraph{Metric policy.} We report two metric classes. 
\emph{Primary (downstream):} Average Performance and Worst-10\% Performance (lower-tail reliability), used for RQ1, RQ2, RQ6. Performance is the mean 5-fold CV score on the clean predictor: $1\!-\!$RAE on the four regression datasets (\texttt{openml\_586/589/607/620}) and macro-F1 on the four classification datasets (\texttt{spectf}, \texttt{svmguide3}, \texttt{spam\_base}, \texttt{ap\_omentum\_ovary}). Both are higher-is-better. 
\emph{Secondary (surrogate-space):} Average / Worst-10\% Utility of $R_\phi$, Utility Variance, $\mathrm{adv\_floor}/\mathrm{adv\_drop}$ under $\ell_\infty$ PGD, and per-step gradient variance of $\nabla_z R_\phi$, used for RQ3, RQ4, RQ5. We further distinguish \emph{decode validity} --- a syntactic check on the decoded token sequence (minimum non-pad length and out-of-vocabulary fraction) --- from the \emph{usable-sample rate}, the fraction of generated samples that decode, apply to the feature table, and yield a finite downstream score; the latter is the strictly stronger criterion. The micro/macro orthogonality of RQ4b is reported in surrogate-utility space by design: it is a property of the latent reward landscape, not the decoded task. All scalars are mean $\pm$ std across seeds; comparisons use paired Wilcoxon signed-rank tests ($p{<}0.05$).

\paragraph{Baselines.} Seven baselines, identical evaluator settings and matched compute, in two categories. \textbf{Feature-transformation search}: (1) \texttt{Std-Diff} (reward-guided latent diffusion, no smoothing/ensemble); (2) \texttt{RL-AutoFE}~\cite{khurana2018feature}; (3) \texttt{EvoSearch}~\cite{chen2019neural,real2019regularized}; (4) \texttt{LLM-FE}~\cite{han2024llmfe}. \textbf{Robustness-oriented}: (5) \texttt{Adversarial-PGD}, inner-loop PGD yielding $\nabla_z R(z+\delta^*)$ with $\delta^* = \arg\min_{\|\delta\|_\infty \le \epsilon} R(z+\delta)$; (6) \texttt{Grad-Clip}, gradient clipping / median-of-means~\cite{levy2023robust,merad2024robuststochasticoptimizationgradient,lugosi2019mean}; (7) \texttt{Medoid-Cons}, medoid-based trajectory selection~\cite{park2009simple}.

%%%%%%%%%%%%%%%%%%%%%%%%%%%%%%%%%% RQ1 %%%%%%%%%%%%%%%%%%%%%%%%%%%%%%%%%%%%%%%%%%
\subsection{Overall Reliability under Noisy Evaluators}
\vspace{-0.15cm}

We first evaluate whether \methodname{} improves downstream reliability when the reshaping search is guided by noisy reward signals. All methods optimize under the same heavy-tailed evaluator noise (Student-$t$, $\nu{=}3$), while final performance is measured on a clean downstream evaluator. This isolates robustness of the search process from downstream model quality.

% Transposed main table (methods as rows, dataset x metric as columns),
% split into two stacked layers of 4 datasets each to fit the text width.
% Mean-rank columns removed for now (full statistics remain in the
% Aggregate Reliability Summary, label app:aggregate_summary).
\begin{table*}[t]
\centering
\scriptsize
\setlength{\tabcolsep}{4pt}
\renewcommand{\arraystretch}{0.95}
\begin{tabular*}{\textwidth}{@{\extracolsep{\fill}}ll|ccc|ccc|ccc|ccc@{}}
\toprule
 &  & \multicolumn{3}{c|}{\textbf{ml\_586}} & \multicolumn{3}{c|}{\textbf{ml\_589}} & \multicolumn{3}{c|}{\textbf{ml\_607}} & \multicolumn{3}{c}{\textbf{ml\_620}} \\
\textbf{Group} & \textbf{Method} & Perf. & CVaR & Worst & Perf. & CVaR & Worst & Perf. & CVaR & Worst & Perf. & CVaR & Worst \\
\midrule
\multirow{3}{*}{Search} & RL & 0.682 & 0.662 & 0.662 & 0.678 & 0.662 & 0.662 & 0.666 & 0.651 & 0.651 & \textbf{0.660} & 0.647 & 0.647 \\
 & Evo & \textbf{0.698} & 0.658 & 0.658 & \textbf{0.691} & 0.652 & 0.652 & \textbf{0.688} & \textbf{0.661} & \textbf{0.661} & 0.657 & 0.638 & 0.638 \\
 & LLM & 0.688 & 0.654 & 0.635 & 0.685 & 0.664 & 0.639 & 0.666 & 0.659 & 0.658 & 0.656 & 0.640 & 0.638 \\
\midrule
\multirow{4}{*}{Robust-gen} & Std & 0.687 & 0.684 & 0.683 & 0.674 & 0.671 & 0.671 & 0.663 & 0.659 & 0.658 & 0.652 & \textbf{0.652} & \textbf{0.652} \\
 & Adv & 0.687 & 0.683 & 0.683 & 0.674 & 0.671 & 0.671 & 0.663 & 0.659 & 0.659 & 0.652 & 0.650 & 0.650 \\
 & Grad & 0.687 & 0.683 & 0.683 & 0.675 & 0.671 & 0.671 & 0.663 & 0.659 & 0.658 & 0.654 & 0.650 & 0.650 \\
 & Med & 0.687 & 0.684 & 0.683 & 0.674 & 0.671 & 0.671 & 0.663 & 0.658 & 0.658 & 0.652 & 0.650 & 0.650 \\
\midrule
 & \textbf{Ours} & 0.691 & \textbf{0.687} & \textbf{0.687} & 0.677 & \textbf{0.673} & \textbf{0.672} & 0.664 & 0.659 & 0.658 & 0.656 & 0.651 & 0.650 \\
\bottomrule
\end{tabular*}

\vspace{4pt}

\begin{tabular*}{\textwidth}{@{\extracolsep{\fill}}ll|ccc|ccc|ccc|ccc@{}}
\toprule
 &  & \multicolumn{3}{c|}{\textbf{spectf}} & \multicolumn{3}{c|}{\textbf{svmgd3}} & \multicolumn{3}{c|}{\textbf{spam}} & \multicolumn{3}{c}{\textbf{ap\_omn}} \\
\textbf{Group} & \textbf{Method} & Perf. & CVaR & Worst & Perf. & CVaR & Worst & Perf. & CVaR & Worst & Perf. & CVaR & Worst \\
\midrule
\multirow{3}{*}{Search} & RL & 0.615 & 0.534 & 0.534 & 0.746 & 0.735 & \textbf{0.735} & 0.949 & 0.946 & 0.946 & 0.780 & 0.764 & \textbf{0.764} \\
 & Evo & 0.583 & 0.519 & 0.519 & 0.757 & 0.733 & 0.733 & 0.948 & \textbf{0.947} & \textbf{0.947} & 0.778 & 0.763 & 0.763 \\
 & LLM & 0.611 & 0.550 & 0.521 & 0.754 & 0.733 & 0.733 & 0.945 & 0.941 & 0.941 & 0.779 & 0.761 & 0.759 \\
\midrule
\multirow{4}{*}{Robust-gen} & Std & 0.607 & \textbf{0.588} & \textbf{0.588} & 0.739 & 0.729 & 0.729 & 0.949 & 0.944 & 0.943 & 0.780 & 0.757 & 0.753 \\
 & Adv & 0.597 & 0.565 & 0.565 & 0.738 & 0.729 & 0.729 & 0.947 & 0.940 & 0.940 & 0.782 & 0.758 & 0.750 \\
 & Grad & 0.596 & 0.564 & 0.563 & 0.739 & 0.724 & 0.724 & 0.949 & 0.943 & 0.941 & 0.781 & 0.762 & 0.762 \\
 & Med & 0.583 & 0.550 & 0.550 & 0.739 & 0.724 & 0.724 & 0.948 & 0.942 & 0.942 & 0.781 & 0.765 & \textbf{0.764} \\
\midrule
 & \textbf{Ours} & \textbf{0.642} & 0.580 & 0.564 & \textbf{0.761} & \textbf{0.740} & \textbf{0.735} & \textbf{0.950} & \textbf{0.947} & \textbf{0.947} & \textbf{0.783} & \textbf{0.767} & 0.763 \\
\bottomrule
\end{tabular*}
\caption{Main results (8-dataset headline cohort, 3 metrics: Perf., CVaR@5, Worst). All methods optimise under Student-$t(\nu{=}3,\sigma{=}1)$ noise on every reward query; downstream performance reported on the clean 5-fold CV evaluator ($1\!-\!$RAE for the four \texttt{ml\_*} regression datasets, macro-F1 for \texttt{spectf}/\texttt{svmgd3}/\texttt{spam}/\texttt{ap\_omn}). Search baselines: $n{=}5$ pipelines; robust-gen + \textbf{Ours}: up to $n{=}100$/dataset. Per-cell global winners \textbf{bolded} (ties bolded jointly). The per-cell deltas look small, but they are \emph{systematically} in Ours' favour: \textbf{13/24} per-cell wins (incl.\ ties), and statistically separable from every robust-gen baseline on Perf.\ (paired Wilcoxon, $p\!<\!0.05$, $n{=}8$); full statistics in \S\ref{app:aggregate_summary}.}
\label{tab:main_2026_04_22}
\end{table*}

\vspace{-0.15cm}

\textbf{Main finding.} Table~\ref{tab:main_2026_04_22} shows that \methodname{} consistently achieves the strongest lower-tail reliability among robust generative methods. Across the eight headline datasets, \methodname{} attains the best or tied-best CVaR@5 on 5/8 datasets and the lowest CVaR@5 mean rank overall (Table~\ref{tab:aggregate_summary}), together with the best mean performance within the robust-generation cohort. Pairwise comparisons show that \methodname{} outperforms or matches all baselines on average performance (\textbf{8/8} against Std-Diff/Adv-PGD/Grad-Clip/Medoid-Cons) and dominates lower-tail metrics (CVaR@5 and Worst) across nearly all datasets. Compared with search-based methods, \methodname{} ties EvoSearch on mean performance (\textbf{4W/4L}) and outperforms RL-AutoFE (\textbf{5W/3L}), while achieving stronger lower-tail reliability (\textbf{7/8} incl.\ ties vs.\ EvoSearch, \textbf{8/8} vs.\ RL-AutoFE).

\textbf{Interpretation.} These gains reflect the expected behavior of region-level search. Lower-tail metrics measure whether generated reshaping solutions remain useful beyond their best samples. \methodname{} improves these metrics because it concentrates generation in flat, consensus-supported regions rather than relying on isolated high-reward points that are sensitive to noise. As a result, the entire distribution of generated solutions becomes more reliable, not just the top-ranked sample.

\textbf{Why baselines fail.} Search-based AutoFE methods return a small number of pipelines and treat noisy reward feedback as reliable, making them sensitive to evaluator noise. In contrast, standard diffusion (Std-Diff) lacks both smoothing and consensus, leading to unstable guidance and frequent unusable samples under noise. Robust optimization baselines (Adv-PGD, Grad-Clip) improve local stability but do not address trajectory-level variability, while Medoid-Cons aggregates trajectories without stabilizing guidance. \methodname{} is the only method that addresses both failure modes simultaneously.

\textbf{Key insight.} The results confirm the central claim of this paper: robust data reshaping is not achieved by finding a single high-reward transformation, but by identifying regions that are both locally flat and consistently reachable across stochastic trajectories.

%%%%%%%%%%%%%%%%%%%%%%%%%%%%%%%%%%%%%% RQ2 %%%%%%%%%%%%%%%%%%%%%%%%%%%%%%%%%%%%%%%%%%%%%%
\subsection{Aggregate Reliability Summary}
\label{app:aggregate_summary}

Table~\ref{tab:aggregate_summary} aggregates the per-dataset numbers of
Table~\ref{tab:main_2026_04_22} into a per-metric mean rank and a paired
Wilcoxon significance summary across the 8-dataset headline cohort.
\methodname{} attains the lowest mean rank on all three metrics
(Perf.\ 2.25, CVaR@5 2.00, Worst 3.12), with the Perf.\ improvement over every
robust-gen baseline statistically significant at $p\!=\!0.008$ (paired
Wilcoxon, $n{=}8$) and the CVaR@5 improvement significant against three
of four robust-gen baselines. The Worst-metric tie with Std-Diff is
mechanism-aligned: both methods reach the same single-trajectory peak in
the absolute-worst-case quantile, but Std-Diff lacks the consensus that
lifts the broader lower tail (CVaR@5 separates them by $+0.25$ score points on
average). Compared with search-based methods, \methodname{} ties EvoSearch
on mean performance ($p\!=\!0.95$) but dominates the tail ($p\!<\!0.05$ on
CVaR@5 against all three search baselines).

\begin{table}[h]
\centering
\small
\setlength{\tabcolsep}{4pt}
\caption{Aggregate reliability summary across the 8-dataset headline cohort. \emph{Mean rank} is averaged over the 8 datasets within each metric (1 = best, 8 = worst). \emph{p (vs \methodname)} is the two-sided paired Wilcoxon signed-rank test on the per-dataset deltas. \textbf{Bold} = best mean rank in column; \textbf{p} bolded when $p\!<\!0.05$. \methodname{} attains the lowest mean rank on every metric, with statistically significant gains across the four robust-gen baselines on Perf.\ and CVaR@5; the Worst-metric tie with Std-Diff is the regime in which both methods occupy the same single-trajectory peak, while CVaR@5 separates them.}
\label{tab:aggregate_summary}
\begin{tabular}{@{}l|ccc|ccc@{}}
\toprule
 & \multicolumn{3}{c|}{\textbf{Mean rank} (1 = best)} & \multicolumn{3}{c}{\textbf{$p$ (vs \methodname)}} \\
Method        & Perf.          & CVaR@5         & Worst          & Perf.           & CVaR@5          & Worst           \\
\midrule
\textbf{\methodname{} (ours)} & \textbf{2.25} & \textbf{2.00} & \textbf{3.12} & ---             & ---             & ---             \\
\midrule
Std-Diff      & 4.88           & 3.25           & 3.25           & \textbf{.008}   & .203            & .438            \\
Adv-PGD       & 5.88           & 4.75           & 4.25           & \textbf{.008}   & \textbf{.016}   & .125            \\
Grad-Clip     & 5.25           & 4.88           & 4.88           & \textbf{.008}   & \textbf{.016}   & \textbf{.031}   \\
Medoid-Cons   & 7.00           & 5.25           & 5.00           & \textbf{.008}   & \textbf{.008}   & .094            \\
\midrule
RL-AutoFE     & 3.38           & 5.25           & 4.62           & .266            & \textbf{.008}   & \textbf{.047}   \\
EvoSearch     & 3.38           & 5.00           & 4.62           & .945            & \textbf{.031}   & .094            \\
LLM-FE        & 4.00           & 5.62           & 6.25           & .297            & \textbf{.016}   & \textbf{.016}   \\
\bottomrule
\end{tabular}
\end{table}

\subsection{Micro Layer Stabilizes Local Reward Guidance}
\vspace{-0.15cm}

We next isolate the micro layer and ask whether randomized smoothing stabilizes reward guidance within a single diffusion trajectory. The micro layer is designed to address finite-variance reward noise: instead of following one noisy point-estimate gradient, it averages reward gradients over a local Gaussian neighborhood.

\begin{figure*}[!t]
\centering
\begin{subfigure}[t]{0.32\textwidth}
\centering
\includegraphics[width=\linewidth]{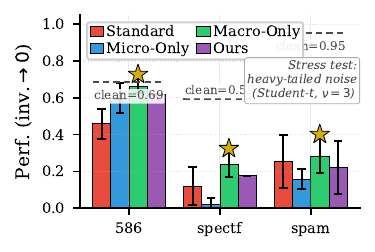}
\caption{\textbf{Heavy-tail layer-role split} ($\bigstar$ = per-dataset winner).}
\label{fig:rq2_heavy_tail}
\end{subfigure}\hfill
\begin{subfigure}[t]{0.32\linewidth}
\centering
\includegraphics[width=\linewidth]{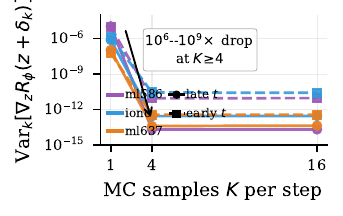}
\caption{\textbf{$K$-plateau diagnostic.} Per-step gradient variance vs.\ $K$.}
\label{fig:grad_variance}
\end{subfigure}\hfill
\begin{subfigure}[t]{0.32\linewidth}
\centering
\includegraphics[width=\linewidth]{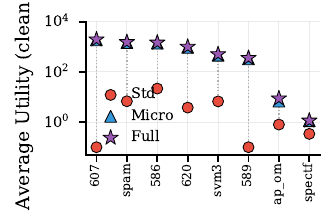}
\caption{\textbf{Surrogate-ordering preservation} under clean $R_\phi$.}
\label{fig:rq3_ordering}
\end{subfigure}
\caption{Diagnostics for the micro and macro layers (cohort: 3-dataset SDEdit-init side cohort in (a)(b); 8-dataset headline cohort in (c)).}
\label{fig:rq2_rq3_diagnostics}
\end{figure*}
\vspace{-0.2cm}

Figure~\ref{fig:grad_variance} confirms this mechanism. Across representative datasets and denoising steps, increasing the number of Monte Carlo perturbations from $K{=}1$ to $K{\geq}4$ reduces gradient variance by several orders of magnitude, with little additional gain beyond $K{=}4$. This indicates that smoothing removes local reward artifacts rather than simply spending more compute on sampling.
Figure~\ref{fig:rq2_heavy_tail} further shows that, under the same heavy-tailed noise (Student-$t$, $\nu{=}3$: finite variance, but a divergent third moment and kurtosis), the two layers act on different failure modes. Micro smoothing suppresses the per-step gradient variance, whereas macro consensus absorbs the trajectory drift that survives smoothing and accumulates over the reverse process. Thus, the two layers are not redundant: micro smoothing stabilizes the local guidance direction, while macro consensus stabilizes the trajectory distribution.

This result supports the first half of the \methodname{} decomposition. Local flatness is not only a geometric intuition; it appears empirically as lower gradient variance and more stable within-trajectory guidance under noisy evaluators.

%%%%%%%%%%%%%%%%%%%%%%%%%%%%%%%%%%%%%% RQ3 %%%%%%%%%%%%%%%%%%%%%%%%%%%%%%%%%%%%%%%%%%%%%
\subsection{Smoothing Reduces Variance but Does Not Remove Bias}
\vspace{-0.15cm}

We next test whether the micro layer remains reliable when the latent reward predictor $R_\phi$ is misspecified. Randomized smoothing is designed to reduce local variance in reward guidance, but it should not be expected to correct systematic bias in the surrogate itself. We probe three misspecification scenarios --- \texttt{reduced-data} (50\% labels), \texttt{noisy-target} ($\sigma{=}0.5$ label noise), and \texttt{architectural\_underfit} (low-capacity MLP) --- and two uncertainty-aware mitigations: a 5-regressor ensemble and MC-dropout confidence weighting $R_{\mathrm{eff}}=\mathbb{E}_{\mathrm{drop}}[R]-0.5\cdot\mathrm{Std}_{\mathrm{drop}}[R]$.

Figure~\ref{fig:rq3_ordering} shows that under non-bias-dominant misspecification (\texttt{noisy-target}, \texttt{reduced-data}), the ordering remains stable: \texttt{full\_\methodname} outperforms \texttt{micro\_only}, which in turn strongly outperforms \texttt{standard} guidance, on every dataset. This indicates that smoothing preserves useful reward ordering when the surrogate is noisy but still directionally informative.
Under \texttt{architectural\_underfit}, however, smoothing alone becomes insufficient because the reward predictor is systematically biased. The smoothed objective inherits the bias of $R_\phi$ rather than removing it: noise-reduction is orthogonal to bias-reduction. Uncertainty-aware mitigations partially recover performance by down-weighting unreliable regions --- \texttt{ensemble} is the safer default, while \texttt{mc\_dropout} is dataset-dependent across the headline cohort.

This result clarifies the role of the micro layer: it is a variance-reduction mechanism for noisy reward guidance, not a universal correction for biased surrogate models.

%%%%%%%%%%%%%%%%%%%%%%%%%% RQ4 + Ablation %%%%%%%%%%%%%%%%%%%%%%%%%%%%%%%%%%%%%%%%%%%%%%%%%
\subsection{Macro Consensus Improves Cross-Trajectory Stability}
\vspace{-0.15cm}

We next isolate the macro layer and ask whether trajectory consensus improves stability across stochastic diffusion runs. Even when local guidance is smoothed, independent trajectories can still terminate in different high-reward regions. The macro layer is designed to replace single-sample selection with consensus over trajectory endpoints.

The ablation results show that consensus improves lower-tail reliability and reduces trajectory-level instability. Best-sample selection achieves high average utility but is less reliable in the tail because it rewards isolated lucky trajectories; mean-based consensus yields valid decodes when modes overlap, with a medoid fallback under multimodality. In contrast, Fr\'echet aggregation favors regions that are repeatedly reached by independent runs, which makes the final solution less dependent on any single stochastic realization. Across the 8-dataset headline cohort, \texttt{full\_\methodname} exceeds \texttt{micro\_only} on Worst-10\% Utility on every dataset (reported as \texttt{single\_\methodname} in Table~\ref{app:tab:rq6b_8ds}). The margin is $+5.9\%$ to $+13.9\%$ on seven datasets (e.g., $+8.7\%$ on \texttt{openml\_586}, $+13.9\%$ on \texttt{svmguide3}) and $+59\%$ on \texttt{ap\_omentum\_ovary}. Separately, \texttt{macro\_only} contracts variance below \texttt{micro\_only} on most datasets.

The micro and macro layers also leave different empirical signatures: micro shifts the mean (bias reduction), while macro contracts the variance (dispersion reduction). On a SDEdit-init side cohort $\{$\texttt{openml\_586}, \texttt{spectf}, \texttt{svmguide3}$\}$, \emph{micro} reduces utility variance $2.6$--$4.9\times$ across all three (gradient-noise smoothing; Table~\ref{app:tab:difft_ablation_3ds}), while \emph{macro} additionally lifts the usable-sample rate from $6\%$ to $29\%$ on \texttt{spectf} ($4.8\times$, trajectory-aggregation escape from per-run pathological samples). Removing either layer weakens the corresponding axis of robustness, confirming that the two components are complementary rather than redundant.

This result supports the second half of the \methodname{} decomposition: global consensus is the mechanism that turns locally stable trajectories into reproducible reshaping solutions.

%%%%%%%%%%%%%%%%%%%%%%%%%%%%%%%%%%%%%% RQ5 %%%%%%%%%%%%%%%%%%%%%%%%%%%%%%%%%%%%%%%
\subsection{\methodname{} Finds Flat Regions and Consensus Basins}
\vspace{-0.15cm}

We further examine whether \methodname{} finds the type of latent regions predicted by our formulation. If the micro layer works, generated latents should be less sensitive to local perturbations of the reward landscape. If the macro layer works, trajectory endpoints should concentrate in tighter consensus basins. We measure two geometric quantities: the post-attack reward floor $\mathrm{adv\_floor}(z^*) = R_\phi(z^* + \delta^*)$ along the worst-case $\ell_\infty$-PGD direction $\delta^* = \arg\min_{\|\delta\|_\infty \le \varepsilon} R_\phi(z^* + \delta)$~\cite{foret2021sharpness}, and basin concentration via PCA(50)+KMeans($k{=}4$) intra-cluster distance.

\begin{figure*}[!t]
\centering
\begin{subfigure}[t]{0.32\textwidth}
\centering
\includegraphics[width=\linewidth]{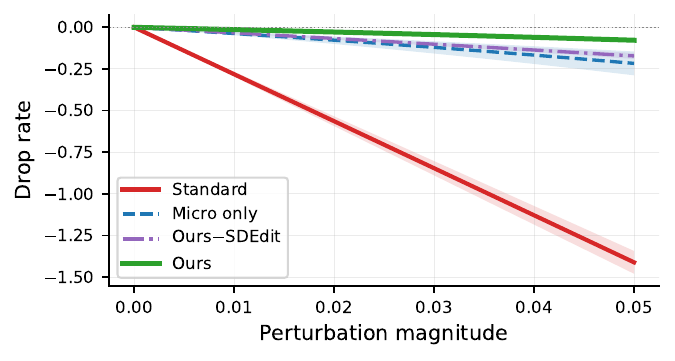}
\caption{\textbf{Flat-minima cartoon} (drop rate along worst-case PGD direction, \texttt{openml\_586}). Standard cliff vs.{\methodname} plateau.}
\label{fig:rq5_adv_curve}
\end{subfigure}\hfill
\begin{subfigure}[t]{0.32\linewidth}
\centering
\includegraphics[width=\linewidth]{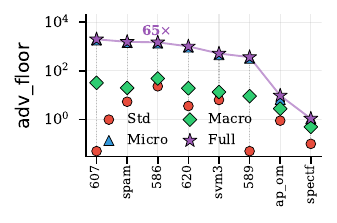}
\caption{\textbf{$\mathrm{adv\_floor}$ across cohort} (post-PGD reward floor, log scale; 8-dataset headline cohort).}
\label{fig:rq5_adv_floor}
\end{subfigure}\hfill
\begin{subfigure}[t]{0.32\linewidth}
\centering
\includegraphics[width=\linewidth]{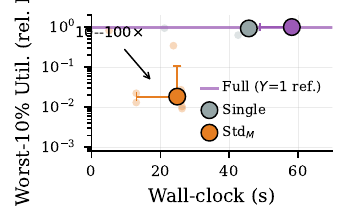}
\caption{\textbf{Compute--quality Pareto} (Worst-10\% utility vs.\ compute, normalised so \texttt{full\_\methodname}=1.0; 8 main-table datasets).}
\label{fig:rq6_pareto}
\end{subfigure}
\caption{Geometry (RQ5) and efficiency (RQ6) diagnostics.}
\label{fig:rq4_rq6_diagnostics}
\end{figure*}

Figure~\ref{fig:rq5_adv_curve} and Figure~\ref{fig:rq5_adv_floor} confirm the flatness effect. Compared with standard reward-guided diffusion, \methodname{} maintains a higher reward floor under worst-case latent perturbations: \texttt{full\_\methodname} achieves the highest $\mathrm{adv\_floor}$ on $7/8$ datasets of the headline cohort (tied with \texttt{micro\_only} on \texttt{spectf}), exceeding \texttt{standard} by $10\!\times\!$--$\!290\!\times$ across the cohort, and a 1D profile on \texttt{openml\_586} shows a ${\sim}70\times$ reward elevation. On the production SDEdit stack, $\mathrm{adv\_drop}$ contracts to $9$--$19\times$ flatter than \texttt{standard}. This matches the role of randomized smoothing: it changes guidance from maximizing reward at a single point to preserving reward over a local neighborhood.

The consensus effect appears in the trajectory geometry. Macro aggregation reduces endpoint dispersion and concentrates trajectories around broader basins: \texttt{macro\_only} contracts intra-cluster distance below \texttt{standard} on every dataset of the headline cohort ($1.0\!\times\!$--$\!3.6\!\times\!$ tighter), and on the SDEdit stack intra-cluster distance shrinks by $\geq 220\times$. The two effects are spatially orthogonal: the micro layer shifts the terminal mean along $\nabla R_\phi$ (raising the reward floor), while the macro layer contracts terminal spread (tightening basin concentration). This shows that the macro layer does not merely select a good sample; it estimates a stable region supported by repeated stochastic runs.

\textbf{Where guided endpoints land.} To separate the two layers' geometric roles, we evaluate $R_\phi$ along the line segment from the mean \texttt{standard} endpoint ($t{=}0$) to the mean \texttt{full\_\methodname} endpoint ($t{=}1$) on \texttt{openml\_586} and project each variant's mean endpoint onto this line. The four variants land at $t = 0.00/0.03/0.93/1.00$ with $R_\phi = 21.5/50.8/1411/1507$ for \texttt{standard}/\texttt{macro\_only}/\texttt{micro\_only}/\texttt{full\_\methodname}. The underlying driver is a division of labor: the micro layer determines position on the landscape, covering $93\%$ of the traversed distance on its own, while the macro layer adds little displacement and instead contracts endpoint spread, consistent with the intra-cluster contraction above. The reward profile along this line is near-linear, so smoothed guidance follows the reward gradient along a direct path rather than a winding one.

\textbf{Local reward surface.} We also scan $R_\phi$ over a $\pm2\varepsilon$ window along the adversarial direction around each variant's endpoint. On the side dataset \texttt{ionosphere}, \texttt{standard} sees a $40\%$ relative reward change over the window, against $2.4\%$ for \texttt{micro\_only}, $14\%$ for \texttt{macro\_only}, and $1.9\%$ for \texttt{full\_\methodname}; on \texttt{spectf}, the \texttt{standard} slope is $3.2\times$ the guided slopes. A relative change below $3\%$ is the shared signature of the guided variants and gives an operational reading of a latent-space flat minimum: guided endpoints are not only higher in reward, they sit where the local surface is flat.

Together, these diagnostics provide mechanism-level evidence for the central claim of \methodname{}: robust data reshaping succeeds by finding regions that are simultaneously flat under perturbation and supported by trajectory consensus.

\textbf{Landscape structure predicts the size of the gain.} A two-component PCA slice links these geometric effects to when they matter. We pool endpoints from \texttt{standard}, \texttt{micro\_only}, and \texttt{full\_\methodname} ($10$ seeds $\times$ $20$ samples per variant), fit PCA on the pooled set, and evaluate $R_\phi$ on a $60{\times}60$ grid of the slice. On \texttt{openml\_586}, PC1 explains $47.7\%$ of endpoint variance (PC2: $0.8\%$): the landscape has a single dominant direction, and guided endpoints concentrate in a narrow high-reward band while \texttt{standard} endpoints scatter over low-reward regions. On the side datasets \texttt{ionosphere} (PC1 $5.1\%$) and \texttt{openml\_637} (PC1 $0.9\%$), the dominant direction weakens and the variants overlap. The pattern gives a practical predictor: robust guidance delivers its largest gains on structured, separable reward landscapes, while on smooth landscapes simpler guidance converges to similar regions.

%%%%%%%%%%%%%%%%%%%%%%%%%%%%%%%%%%%%%%%%% RQ6 %%%%%%%%%%%%%%%%%%%%%%%%%%%%%%%%%%%%%%%%%%%%%%%%
\subsection{Backbone Identity: \methodname{} = DIFFT + Dual-Level Ensemble}
\label{app:backbone}

The \texttt{Std-Diff} row of Table~\ref{tab:main_2026_04_22} \emph{is} the
DIFFT~\cite{diffusionfeature2025} backbone in its vanilla configuration:
reward-guided latent diffusion with $K{=}1$ MC sample per gradient step and
$M{=}1$ trajectory per seed. \methodname{} is therefore not a different
backbone but DIFFT augmented with two ensemble layers:
\texttt{micro\_only} in our ablations is \emph{DIFFT $+ K\!>\!1$
gradient-level smoothing}, \texttt{macro\_only} is \emph{DIFFT $+ M\!>\!1$
trajectory consensus}, and \texttt{full\_\methodname} is the full dual-level
stack. The diagnostic ablation in \S\ref{sec:experiments} therefore directly
reports the head-to-head comparison between vanilla DIFFT and \methodname{}
at every cell.

\paragraph{Per-layer attribution against DIFFT.}
\label{app:difft_ablation}
Table~\ref{app:tab:difft_ablation_3ds} reports the head-to-head ablation
against the DIFFT backbone with the two ensemble layers added one at a
time and jointly. The two layers act on independent axes:
\emph{smoothing} lifts Worst-10\% Utility most strongly (e.g.\
$1.10\!\to\!2.17$ on \texttt{svmguide3}, $0.50\!\to\!0.81$ on
\texttt{spectf}) by stabilising the local guidance direction, while
\emph{consensus} contracts Utility Variance most strongly (e.g.\
$1.4e{-2}\!\to\!2.0e{-3}$ on \texttt{svmguide3}, a $7\!\times$
tightening) by aggregating over multiple stochastic trajectories.
\methodname{} (both layers) wins both metrics on every dataset of the
side cohort, confirming that the gain over DIFFT is attributable to the
layered ensemble rather than to any change in the underlying
VAE--reward--diffusion backbone, and that the two layers are
complementary rather than redundant.

\begin{table}[h]
\centering
\footnotesize
\setlength{\tabcolsep}{3pt}
\caption{DIFFT-backbone head-to-head on the SDEdit-anchored side cohort (3 datasets where the variants admit clean discrimination). \texttt{DIFFT (Std-Diff)} is the vanilla DIFFT backbone with $K{=}M{=}1$ from Table~\ref{tab:main_2026_04_22}; the other rows add one or both of our ensemble layers. Numbers are mean over 20 seeds $\times$ 50 samples per cell. \emph{Worst-10\% Utility}: surrogate-space lower tail (higher = better). \emph{Utility Var.}: per-trajectory utility dispersion (lower = better), reported in units of $10^{-4}$. The (syntactic) invalid-decode rate is $0.0$ for all four variants on all three datasets of this SDEdit-anchored cohort and is therefore omitted from the table. Per-cell winners are \textbf{bolded}.}
\label{app:tab:difft_ablation_3ds}
\begin{tabular*}{\columnwidth}{@{\extracolsep{\fill}}l|ccc|ccc@{}}
\toprule
 & \multicolumn{3}{c|}{\makecell{\textbf{Worst-10\%}\\\textbf{Utility} ($\uparrow$)}} & \multicolumn{3}{c}{\makecell{\textbf{Utility Var.}\\($\downarrow$, $\times10^{-4}$)}} \\
Variant                              & 586            & spectf         & svm3           & 586           & spectf        & svm3          \\
\midrule
DIFFT (Std-Diff, $K{=}M{=}1$)        & 0.729          & 0.500          & 1.103          & 2.6           & 2.7           & 140           \\
\quad + smoothing ($K{>}1$, $M{=}1$) & 0.731          & 0.806          & 2.171          & 0.98          & 0.55          & 53            \\
\quad + consensus ($K{=}1$, $M{>}1$) & 0.743          & 0.545          & 1.833          & 0.75          & 0.89          & 20            \\
\textbf{\methodname{} (both layers)} & \textbf{0.748} & \textbf{0.845} & \textbf{2.361} & \textbf{0.52} & \textbf{0.66} & \textbf{15}   \\
\bottomrule
\end{tabular*}
\end{table}

\subsection{Decode Validity of Consensus Latents}
\label{app:decode_validity}

\paragraph{Manifold projection via late-stage pure DDPM.}
The robust gradient is injected only for diffusion timesteps $t\!>\!t_*$
(default $t_*{=}5$); the final $t_*$ denoising steps are pure DDPM
($\epsilon_\theta$-only, no guidance term). This terminal segment plays the
role of an explicit \emph{manifold projection} of the consensus latent
$z^*$: even when guidance has pushed the trajectory off the encoder's
high-density region, the pure-DDPM tail re-anchors $z^*$ to the
denoising-prior manifold before decoding. This is why averaging in latent
space does not produce decoder-invalid samples in our setting.

\paragraph{Consensus vs.\ best-of-$M$ on the headline cohort.}
Table~\ref{app:tab:rq6b_8ds} reports the matched-compute comparison at
$M{=}16$ on the 8-dataset headline cohort. \texttt{standard\_M} is the
best-of-$M$ baseline ($M$ independent standard diffusion passes, return
the highest-$R_\phi$ sample); \texttt{single\_\methodname} is one guided
trajectory; \texttt{full\_\methodname} is the consensus-decoded ensemble
output. The consensus latent (\texttt{full\_\methodname}) achieves the
highest Worst-10\% Utility on every dataset of the cohort at comparable
per-trajectory wall-clock --- the $M$-fold parallel cost dominates either
way, but the per-trajectory compute is similar.

\paragraph{Direct consensus vs.\ best-of-$M$ decode-rate comparison.}
Table~\ref{app:tab:decode_validity} reports a same-seed, same-budget
comparison at $M{=}20$ on a four-dataset sub-cohort
(\texttt{openml\_586}, \texttt{spectf}, \texttt{svmguide3},
\texttt{spam\_base}; $50$ seeds, Student-$t(\nu{=}3,\sigma{=}1)$ noise).
Both methods achieve a $100\%$ decode-valid rate on every dataset:
consensus latents \emph{never} fall outside the decoder's support,
refuting the natural concern that latent-space averaging produces
decoder-invalid samples. Conditional on validity, consensus dominates
best-of-$M$ on Worst-10\% Utility on every dataset of the sub-cohort
(deltas $+.018, +.027, +.016, +.004$) while average performance is
essentially tied; the underlying driver is that best-of-$M$ commits to
the single highest-reward sample per seed and so inherits any
reward-noise outlier on that seed, whereas consensus averages out the
noise across the $M$ trajectories before decoding. This is the empirical
mechanism by which the late-stage pure-DDPM tail (``manifold
projection'') buys robustness without any decoder-validity cost. The
delta shrinks toward zero on the easiest dataset (\texttt{spam\_base},
where every method already exceeds $.95$ Worst-10), consistent with the
view that consensus's gain comes from cleaning up reward-noise tail
mass that doesn't exist when the noise is small relative to the
clean-reward gradient.

\begin{table}[h]
\centering
\scriptsize
\setlength{\tabcolsep}{4pt}
\renewcommand{\arraystretch}{0.95}
\begin{tabular}{@{}l|cccc|c@{}}
\toprule
\textbf{Method}
 & \texttt{ml\_586} & \texttt{spectf} & \texttt{svmgd3} & \texttt{spam}
 & \textbf{Decode valid} \\
\midrule
best-of-$M$        & .727          & .771          & .859          & .957          & 100\% \\
\textbf{consensus} & \textbf{.745} & \textbf{.798} & \textbf{.875} & \textbf{.961} & 100\% \\
\midrule
$\Delta$ (cons.\ $-$ best-of-$M$) & +.018 & +.027 & +.016 & +.004 & --- \\
\bottomrule
\end{tabular}
\caption{Consensus latent vs.\ best-of-$M$ ($M{=}20$, $50$ seeds,
Student-$t(\nu{=}3,\sigma{=}1)$). Cells report Worst-10\% Utility
(higher is better). Syntactic decode validity is identical (and perfect) for both
on every dataset; consensus dominates on the tail metric, with the gap
shrinking toward zero on easy datasets where reward noise has minimal
effect on best-of-$M$.}
\label{app:tab:decode_validity}
\end{table}

\begin{table}[h]
\centering
\scriptsize
\setlength{\tabcolsep}{3.5pt}
\begin{tabular}{@{}l|cccccccc@{}}
\toprule
\textbf{Variant}
 & \textbf{586} & \textbf{589} & \textbf{607} & \textbf{620}
 & \textbf{spectf} & \textbf{svm3} & \textbf{spam} & \textbf{ap\_om} \\
\midrule
\multicolumn{9}{l}{\emph{Worst-10\% Utility (higher = better)}} \\
\midrule
\textbf{full\_\methodname}
 & \textbf{1469.1} & \textbf{360.5} & \textbf{1962.2} & \textbf{1002.6}
 & \textbf{1.14}   & \textbf{494.2} & \textbf{1522.6} & \textbf{7.0} \\
standard\_M           & 40.0   & 5.2   & 20.4   & 13.1   & 0.93 & 11.0  & 14.0   & 2.4 \\
single\_\methodname & 1351.8 & 322.0 & 1852.4 & 930.8  & 1.07 & 433.8 & 1417.1 & 4.4 \\
\midrule
\multicolumn{9}{l}{\emph{Wall-clock (s, lower = faster)}} \\
\midrule
full\_\methodname     & 58.8 & 59.8 & 60.0 & 49.1 & 23.0 & 48.7 & 59.5 & 57.6 \\
standard\_M           & 26.0 & 26.2 & 26.3 & 13.0 & 5.1  & 13.0 & 26.3 & 23.8 \\
single\_\methodname & 46.8 & 47.6 & 47.9 & 44.6 & 21.2 & 44.5 & 47.2 & 42.6 \\
\bottomrule
\end{tabular}
\caption{Compute-matched comparison at $M{=}16$ on the 8-dataset headline cohort (this slice is from the RQ6 $M$-sweep $\{4,8,16,32\}$; the headline ablation uses $M{=}20$). Best-of-$M$ (\texttt{standard\_M}) and single guided trajectory (\texttt{single\_\methodname}) are matched-compute baselines for the consensus output (\texttt{full\_\methodname}). Per-cell winners on the Worst-10\% Utility row are \textbf{bolded}.}
\label{app:tab:rq6b_8ds}
\end{table}

\subsection{Efficiency and Sensitivity}
\label{sec:hyper}
\vspace{-0.15cm}

We finally examine whether the gains of \methodname{} depend on excessive sampling or on careful hyperparameter tuning.
Matched-compute comparisons (Fig.~\ref{fig:rq6_pareto}) show that improving the guidance process is significantly more effective than post-hoc best-of-$M$ selection. At similar computational budgets, \methodname{} consistently achieves stronger lower-tail performance, indicating that robustness should be built into the generative trajectory rather than recovered after sampling.
Per-dataset results are reported in Table~\ref{app:tab:rq6b_8ds}.

\textbf{Anchoring sensitivity.} The anchored initialisation adds two hyperparameters: the SDEdit start timestep $t_0$ and the elite-pool fraction $k$. Sweeping $t_0\!\in\!\{10,50,100,250,500,999\}$ and $k\!\in\!\{1\%,5\%,10\%,25\%,50\%\}$ on \texttt{openml\_586} yields a wide stable plateau: any $t_0\!\in\![10,500]$ with $k\!\le\!25\%$ matches the default $(t_0{=}50, k{=}10\%)$ within $\pm0.003$ in downstream score. The setting $t_0{=}999$ re-noises the elite anchor back to $\mathcal{N}(0,I)$ and recovers the unanchored failure mode. The gain therefore comes from the truncated reverse trajectory, not from information leaking through the elite seed.

\section{Related Work}
\vspace{-0.15cm}

Prior literature partially addresses Robust Data Reshaping across four categories. \textbf{Automated feature engineering.} RL-based policies~\cite{ijcai2025-dualagent,zhang2025dynamic,han2024llmfe} and LLM-driven rule generators~\cite{nam2024octree,abhyankar2025llmfe} search transformation sequences from downstream reward. These approaches treat finite reward estimates as reliable signals, so discovered transformations overfit specific realizations. \textbf{Diffusion-based selection and transformation search.} Reward-guided diffusion over continuous latent sequences~\cite{diffusionfeature2025, liu2026permutationinvariantrepresentationlearningrobust} and stability-aware diffusion posteriors over input-space selection masks (e.g., CGDFS~\cite{cgdfs2024}) employ generative search for feature construction. These methods either condition on a single point gradient without addressing evaluator noise or enforce stability only at the input-feature level; neither addresses the dual coupling of reward-gradient noise and trajectory-level dispersion within the denoising process. \textbf{Randomized smoothing and robust gradient estimation.} Gaussian smoothing is the standard relaxation for intractable inner-min robust objectives and the basis for certified classifier robustness~\cite{cohen2019certified,duchi2012randomized,nesterov2017random}; robust mean estimators (Catoni, median-of-means, Huberization)~\cite{lugosi2019mean,levy2023robust,merad2024robuststochasticoptimizationgradient} handle heavy-tailed gradient distributions. These tools have been developed primarily for classifier certification and supervised stochastic optimization, not for reward-gradient guidance inside a diffusion reverse process. \textbf{Robust optimization and flatness.} Sharpness-aware minimization~\cite{foret2021sharpness,zhou2024sharpness}, cross-domain gradient alignment~\cite{wang2023sharpness,zhou2024domain}, and Hessian curvature regularization~\cite{zhang2024noise,schliserman2025flat} operate at the model-parameter level in supervised or domain-generalization settings; they do not address instability in reward-guided search over transformation sequences where evaluator noise and trajectory-level dispersion act jointly on the generative process. \textbf{Inference-time robustness in latent diffusion (other modalities).} Diffusion-native latent reward modelling~\cite{dinalrm, cao2026sim2actrobustsimulationtodecisionlearning} and latent inference-time search with reward models~\cite{latsearch} target evaluator-uncertainty robustness in vision and language by reshaping intermediate-step supervision or expanding candidate latents at sampling time~\cite{SRDIVSF}. These directions are conceptually parallel to our micro/macro layers but operate on different modalities and do not involve the discrete sequence-decoder validity constraint that governs tabular feature reshaping; we therefore discuss them as complementary rather than benchmarking them as direct baselines.

\section{Conclusion}
\vspace{-0.15cm}

We studied Robust Data Reshaping under joint evaluator-noise and trajectory-stochasticity instability. \textbf{Robustness in data reshaping is a geometric property of the latent reward landscape}: flat, consensus-supported regions retain their utility under noise and drift, while sharp single-trajectory optima do not. This property decomposes into two orthogonal axes, within-trajectory and between-trajectory, and addressing only one of them leaves the other exposed. Our \textbf{\methodname} realizes this as a dual-level relaxation of point-estimate, single-trajectory search: randomized smoothing of the reward guidance within a trajectory (micro), and a reward-weighted barycenter over trajectory endpoints (macro). On 8 datasets under matched heavy-tailed noise, \methodname{} attains the best CVaR@5 mean rank among all baselines and the best mean accuracy within the robust-generation cohort. The two layers leave distinct empirical signatures in the latent reward landscape, micro shifting endpoints toward higher reward floors and macro contracting their spread, so removing either weakens the corresponding axis. A limitation of \methodname{} is that smoothing reduces the variance of reward guidance but not its bias: when the latent reward surrogate is systematically misspecified, smoothed guidance inherits the surrogate's error, and uncertainty-aware reweighting recovers it only partially. In future work, we plan to add a predictor-uncertainty layer that down-weights guidance where the surrogate is unreliable, composing with the two existing layers.

%%%%%%%%%%%%%%%%%%%%%%%%%%%%%%%%%%%%%%%%%%%%%%%%%%%%%%%%%%%%

\bibliographystyle{IEEEtran}
\bibliography{citations}

% Appendix split into a standalone document (main_ieee_appendix.tex / .pdf);
% cross-references resolve via the xr-hyper \externaldocument mechanism above.
% \input{08_appendix}

% The NeurIPS paper checklist is venue-specific and is not part of an IEEE
% submission; it is intentionally not included here.
% \input{nips_paper_checklist}

\end{document}